# Machine Learning for Biomedical Raman Spectroscopy: From Spectral Acquisition to Clinical Translation

Bogdan Oancea[1,2], Ana Maria Seciu-Grama[1], Nicoleta Siminea[1], Laura Mihaela Stefan[1], Alice Stoica[1], Joel Sjöberg[4], Marian Necula[1,2], Ana-Maria Prelipcean[1], Corneliu Ovidiu Vrancianu[1,2], Eduard Milea[1], Andrei Păun[2,3], Ion Petre[1,4], Mihaela Păun[1,2]

[1] National Institute of Research and Development for Biological Sciences, Romania
[2] University of Bucharest, Romania
[3] Research Institute for Artificial Intelligence "Mihai Draganescu", Romanian Academy
[4] University of Turku, Finland

# Abstract

Raman spectroscopy provides label-free, chemically specific characterization of biological systems and has emerged as a powerful tool for applications ranging from cancer diagnosis and molecular subtyping to microbiological identification and intraoperative decision support. However, biomedical Raman spectra are high-dimensional, noisy, and strongly affected by fluorescence background, acquisition variability, and biological heterogeneity, creating substantial challenges for computational analysis. Machine learning has become an essential component of modern Raman spectroscopy workflows, enabling extraction of diagnostically relevant information from complex spectral data.

This review examines the role of machine learning across the complete biomedical Raman spectroscopy pipeline. We discuss preprocessing and signal correction, unsupervised structure discovery, supervised learning for diagnosis and molecular stratification, emerging representation-learning and transfer-learning approaches, explainability and Raman biomarker discovery, and multimodal integration with imaging, pathology, and molecular profiling technologies. Particular attention is given to the growing role of machine learning in supporting biologically interpretable and clinically actionable analyses rather than simple diagnostic classification.

We examine the practical challenges that continue to limit clinical translation. These include limited dataset sizes, inter-instrument variability, inconsistent preprocessing practices, insufficient external validation, reproducibility concerns, and barriers to software, data, and metadata sharing. We argue that future progress will depend not only on advances in machine-learning methodology but also on coordinated efforts toward standardization, robust validation, explainability, and deployment-ready analytical frameworks.

By integrating methodological, biomedical, and translational perspectives, this review provides a comprehensive overview of the current state of machine learning in biomedical Raman spectroscopy and outlines key directions for the development of robust and clinically deployable Raman-AI systems.

# Introduction

Raman spectroscopy has become an increasingly important analytical technique in biomedical research, providing label-free, non-destructive, and chemically specific information about biological samples without requiring staining or extensive sample preparation (Butler et al., 2016). Since its first experimental observation nearly a century ago (Raman and Krishnan, 1928), the Raman effect has offered a powerful basis for probing molecular vibrations: the characteristic frequency shifts produced by inelastic light scattering generate biochemical fingerprints that reflect the molecular composition, structural organization, and metabolic state of complex biological materials. In biomedical contexts, Raman spectra simultaneously encode vibrational signatures associated with proteins, lipids, nucleic acids, carbohydrates, and metabolites, thereby capturing molecular alterations related to disease onset, tissue organization, microbial identity, and treatment response (Pence and Mahadevan-Jansen, 2016). These properties have established Raman spectroscopy as a versatile platform for biomedical analysis, supporting applications ranging from cancer diagnosis and molecular subtyping to microbiological identification and intraoperative decision support, spanning applications from single-cell analysis to real-time surgical guidance (Hollon et al., 2023). At the same time, biomedical Raman spectra are high-dimensional, strongly correlated, and simultaneously affected by fluorescence background, detector noise, cosmic ray artefacts, acquisition variability, and biological heterogeneity, a combination of challenges that makes automated computational analysis not merely useful, but necessary (Luo et al., 2022). The field has matured to the point where methodological developments, biomedical applications, and translational challenges must be considered together rather than as independent research directions.

Machine learning has become a central component of modern biomedical Raman spectroscopy because it can extract diagnostically relevant patterns from complex spectral data that are difficult to identify using conventional analytical approaches (Boateng, 2025). Machine-learning methods now support a broad range of tasks, including preprocessing, tissue segmentation, classification, biomarker discovery, and multimodal data integration (Kazemzadeh et al., 2022b, 2022a; Lita et al., 2024). Their ability to model nonlinear relationships and scale to large hyperspectral datasets is increasingly important as Raman spectroscopy moves toward real-time and clinically integrated applications (Kondepudi et al., 2025; Lita et al., 2024). Figure 1 gives a visual conceptual map of the field.

Despite these advances, the translation of machine-learning-assisted Raman spectroscopy into robust and reproducible biomedical practice remains substantially incomplete. Limited dataset sizes, pronounced inter-instrument and inter-laboratory variability, inconsistent preprocessing pipelines, inadequate external validation, and restricted sharing of code and data continue to constrain the generalizability and clinical credibility of reported results (Semmelrock et al., 2025). High classification accuracy alone does not confer clinical utility; rather, models must also be interpretable, biologically plausible, and stable across patient populations and instrument configurations (Huang et al., 2023).

In this review, we examine the methodological foundations, current applications, and practical challenges of machine learning for biomedical Raman spectroscopy. Existing reviews (Boateng, 2025; Butler et al., 2016; Krafft and Popp, 2015; Pence and Mahadevan-Jansen, 2016) have largely focused on Raman instrumentation, spectroscopic methodologies, specific biomedical applications, or comparative evaluation of machine-learning algorithms, whereas less attention has been given to the complete analytical pipeline linking spectral acquisition, machine-learning methodology, biological interpretation, and clinical translation. This review instead focuses on the methodological factors that determine robustness, reproducibility, interpretability, and clinical translatability of machine-learning-based Raman spectroscopy. We discuss the complete analytical pipeline, including preprocessing, clustering, classification, explainability, multimodal integration, and software infrastructure, and conclude by examining the challenges of standardization, reproducibility, generalization, and clinical deployment that continue to limit translation into routine biomedical practice. Throughout the review, we emphasize the progression from spectral acquisition and computational analysis to biological interpretation and clinical deployment, reflecting the increasingly translational role of machine learning in biomedical Raman spectroscopy.

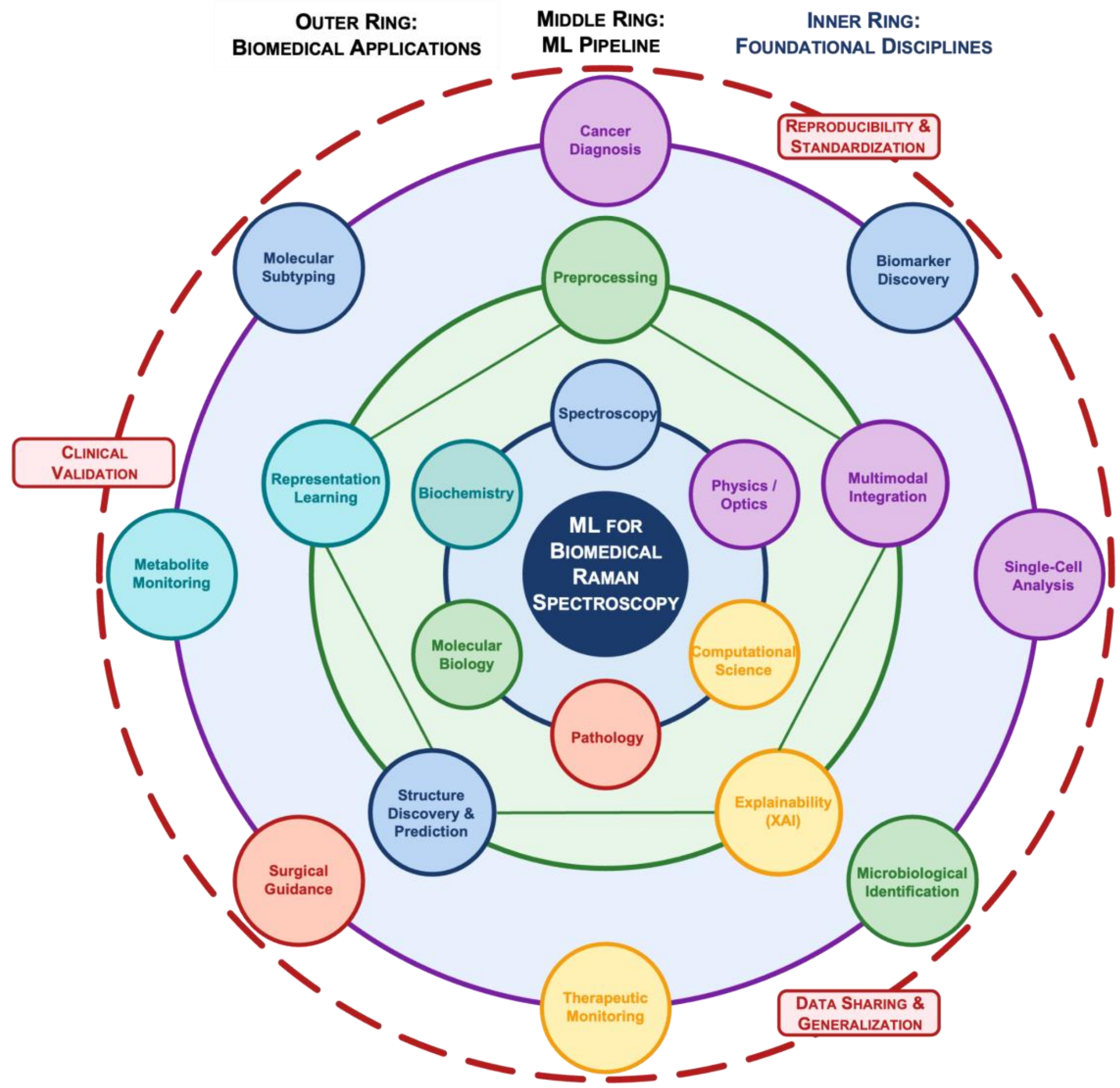

**Figure 1. Conceptual landscape of machine learning for biomedical Raman spectroscopy.** The field integrates foundational disciplines including spectroscopy, physics,

molecular biology, pathology, and computational science with methodological components such as preprocessing, representation learning, predictive modeling, explainability, and multimodal integration. These approaches support a broad range of biomedical applications, including cancer diagnosis, molecular subtyping, microbiological identification, biomarker discovery, surgical guidance, and therapeutic monitoring. Surrounding all stages of the workflow are key translational challenges, including standardization, reproducibility, clinical validation, data sharing, and model generalization, which remain central barriers to routine clinical deployment.

# Physical and Computational Characteristics of Raman Spectroscopy Data

Raman spectroscopy is a vibrational spectroscopic technique based on the inelastic scattering of light by molecules. When monochromatic light interacts with matter, most photons are elastically scattered without energy change (Rayleigh scattering), while a small fraction undergoes energy exchange with molecular vibrational modes, producing Raman scattering. The resulting frequency shifts provide biochemical fingerprints that reflect the molecular composition and structural organization of the sample. First observed almost a century ago (Raman and Krishnan, 1928), the effect provides a powerful spectroscopic tool for investigating vibrational, rotational, and other low-frequency modes in molecules.

Raman spectroscopy is label-free, non-destructive, requires minimal sample preparation, and works in aqueous environments, making it ideal for biological, chemical, and material science applications. The frequency shift provides a molecular fingerprint unique to each substance, enabling identification and structural analysis. Additionally, since the Raman effect is sensitive to molecular symmetry, it complements infrared spectroscopy and allows for the detection of vibrational modes that may be IR-inactive (Smith and Dent, 2004). In biological samples, Raman spectra can capture information related to proteins, lipids, nucleic acids, and metabolites simultaneously, enabling molecular characterization of tissues, cells, and biofluids without staining or extensive sample preparation (Krafft and Popp, 2015).

In biological systems, Raman spectra rarely correspond to isolated molecular compounds. Instead, biomedical spectra typically represent complex superpositions of signals originating from proteins, lipids, nucleic acids, carbohydrates, and metabolites. Additional variability arises from tissue heterogeneity, cellular composition, inflammation, necrosis, and differences in acquisition conditions. Consequently, biomedical Raman spectra often exhibit overlapping peaks, weak discriminative signals, and substantial intra- and inter-patient variability (Movasaghi et al., 2007). A major challenge in biomedical Raman spectroscopy is the inherently weak intensity of Raman scattering, often several orders of magnitude lower than the excitation signal. Biological samples additionally exhibit strong fluorescence background, which can obscure diagnostically relevant Raman peaks. Detector noise, cosmic ray artifacts, and acquisition variability further complicate spectral analysis, motivating the extensive use of preprocessing, chemometric analysis, and machine learning methods (Boateng, 2025; Luo et al., 2022).

Several Raman modalities have been developed to address specific biomedical limitations. Spontaneous Raman spectroscopy provides broad biochemical characterization but is limited by weak signal intensity and relatively slow acquisition speed. Surface-enhanced

Raman spectroscopy (SERS) dramatically increases sensitivity through plasmonic enhancement, enabling detection of low-abundance biomarkers and even single molecules Vázquez-Iglesias et al., 2024). Resonance Raman spectroscopy (RRS) selectively amplifies vibrational modes associated with specific chromophores by matching the excitation wavelength to electronic transitions. Coherent anti-Stokes Raman spectroscopy (CARS) and stimulated Raman scattering (SRS) enable rapid label-free imaging with high spatial resolution, making them attractive for microscopy and intraoperative applications (Cheng and Xie, 2016). Spatially offset Raman spectroscopy (SORS) extends Raman analysis to subsurface and deeper tissue interrogation (Mosca et al., 2021).

Raman spectroscopy has been successfully applied across diverse fields. In biomedical research, Raman spectroscopy has enabled non-invasive cancer diagnosis by distinguishing between healthy and malignant tissues based on molecular composition (Zhang et al., 2022). In microbiology, the integration of Raman spectroscopy with deep learning has enabled rapid label-free identification of pathogenic bacteria from complex spectral profiles, highlighting the potential of AI-assisted Raman analysis for clinical diagnostics (Ho et al., 2019). In materials science, Raman spectroscopy has played a pivotal role in characterizing carbon nanomaterials such as graphene and carbon nanotubes, where peak positions and intensities reveal structural and electronic properties (Ferrari, 2007). In pharmaceutical development, Raman imaging assists in detecting polymorphs and monitoring drug formulation processes (Shah et al., 2023). These examples highlight Raman spectroscopy's versatility, offering label-free, nondestructive, and chemically specific insights across scales, from single molecules to complex systems.

From a computational perspective, Raman spectra constitute high-dimensional and highly correlated data representations. Neighboring wavelengths often exhibit strong statistical dependence, while diagnostically relevant information may be confined to relatively small spectral regions. These properties make Raman spectroscopy particularly suitable for multivariate statistical analysis and machine learning approaches capable of extracting subtle biochemical patterns from complex spectral data (Wold et al., 1987).

However, Raman spectroscopy presents substantial analytical challenges arising from weak signals, fluorescence contamination, spectral overlap, and biological heterogeneity. These challenges have driven the development of sophisticated preprocessing pipelines, statistical learning methods, and machine learning approaches for spectral interpretation. The following sections examine how machine learning addresses these challenges, from signal correction and unsupervised exploration to diagnostic classification, biological interpretation, and clinical deployment.

# Preprocessing and Signal Correction

Preprocessing is a critical stage in Raman spectroscopy pipelines because raw biomedical spectra are strongly affected by fluorescence background, detector noise, cosmic ray artifacts, acquisition variability, and intensity distortions (Butler et al., 2016; Movasaghi et al., 2007; Pence and Mahadevan-Jansen, 2016). These effects can obscure diagnostically relevant biochemical signatures and substantially alter downstream statistical or machine learning analyses. Consequently, preprocessing is not merely a technical correction step, but a major determinant of spectral reproducibility, model robustness, and biological interpretability (Beleites et al., 2013; Ntziouni et al., 2022; Semmelrock et al., 2025). However, preprocessing may also alter diagnostically relevant biochemical information. For example, excessive smoothing can distort narrow Raman peaks, aggressive baseline correction may suppress biologically meaningful low-frequency variations, and normalization procedures can artificially modify relative peak intensities (Beleites et al., 2013; Ntziouni et al., 2022).

The main preprocessing operations in biomedical Raman spectroscopy include denoising, baseline correction, artifact removal, normalization, and scaling (Figure 2). These methods are vital for correcting statistical properties of Raman spectra (Cappel et al., 2010; Fang et al., 2024; Uckert et al., 2019; Zhang et al., 2010). Baseline-signals present as non-linear spectrum vectors which offset peak intensities, often changing peak-intensity ratios. Cosmic rays on the other hand present as sharp, narrow peak-like intensities, lacking relevance for the chemical composition of the scanned material. Noise within spectrum intensities cause sporadic shifts between adjacent wavelengths and can in the worst-case scenario effectively obscure peak signals (Audier et al., 2020; Smulko et al., 2015). The silent region is often removed with the goal of reducing the computational load in methods whose runtimes are tied to the number of wavelengths. The location of the silent region is well-documented and is often removed to reduce the number of irrelevant wavelengths (Plakas et al., 2022; Schultz and Popp, 2025). It is possible for components such as cosmic rays and noise to occur within the silent region, which makes its removal beneficial for reducing erroneous signals with potential to skew statistical estimates (Vardaki et al., 2024). Another critical aspect of these components is that any one of them may be present with considerably high intensity, effectively diminishing any noticeable peaks within the spectrum. Although conceptually distinct, these operations interact strongly and can substantially influence downstream clustering, classification, and explainability analyses. Table 1 summarizes their main purpose and limitations.

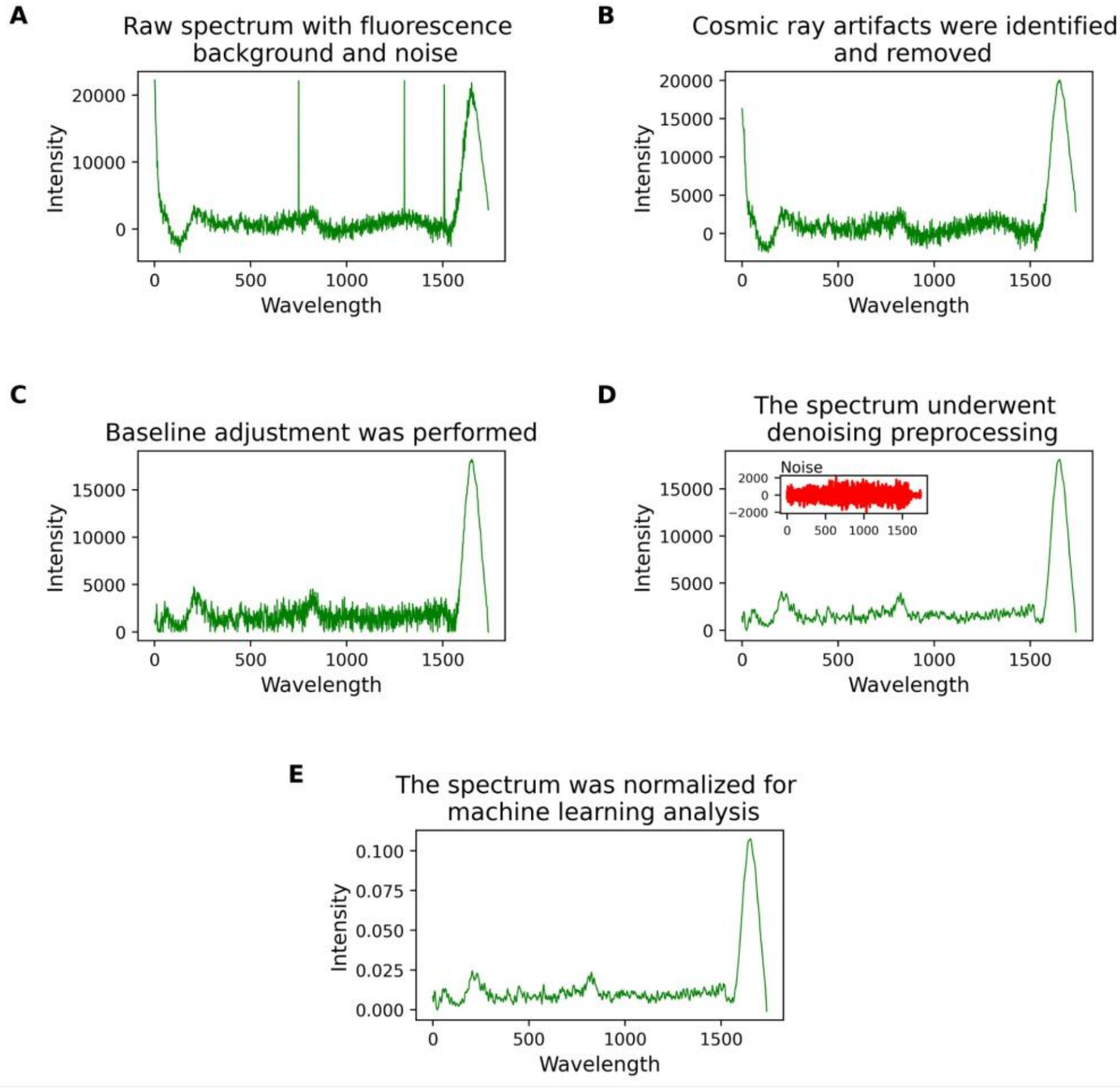

**Figure 2. Representative preprocessing workflow for biomedical Raman spectroscopy.** Raw spectra are sequentially corrected for cosmic-ray artifacts, fluorescence background, and noise before normalization for downstream machine-learning analysis. The figure illustrates how preprocessing progressively improves spectral quality while preserving diagnostically relevant biochemical information.

**Table 1**. Common preprocessing operations in biomedical Raman spectroscopy and their principal objectives and limitations.

| Step | Representative Methods | Main Purpose | Major Risk |
|---|---|---|---|
| Denoising | Savitzky-Golay, Wavelets, Deep Learning | Noise reduction | Peak distortion |
| Baseline correction | ModPoly, airPLS, BubbleFill, EMSC | Fluorescence removal | Loss of biological signal |

| Cosmic ray removal | Thresholding, derivatives, ML methods | Artifact removal | Removal of genuine peaks |
|---|---|---|---|
| Normalization | L2, vector normalization | Reduce intensity variability | Loss of concentration information |
| Scaling | z-score, standard scaling | Improve ML training | Alter feature importance |

## Denoising

Biomedical Raman spectra are frequently affected by detector noise, photon shot noise, fluorescence contamination, and acquisition variability, which may obscure weak biochemical signatures and reduce downstream classification performance. Consequently, denoising constitutes a fundamental preprocessing step aimed at improving spectral fidelity while preserving diagnostically relevant molecular information. Spectral quality is commonly quantified using the signal-to-noise ratio (SNR) (Audier et al., 2020; Liu et al., 2021; Pan et al., 2020; Smulko et al., 2015), which compares the intensity of Raman peaks against estimated background noise levels. A central challenge in Raman denoising is balancing noise suppression against preservation of diagnostically relevant spectral information. Excessive smoothing may distort narrow Raman peaks, alter peak intensity ratios, or suppress weak biomarkers that are critical for downstream classification tasks. This issue is important in biomedical Raman spectroscopy, where subtle spectral variations may correspond to clinically relevant molecular differences.

Savitzky–Golay filtering (Savitzky and Golay, 1964) is one of the most widely used methods in Raman preprocessing. The method applies local polynomial fitting within sliding spectral windows to smooth high-frequency noise while approximately preserving peak morphology. In addition to denoising, Savitzky–Golay filters can also be used for derivative-based peak detection and spectral feature enhancement. The method is highly sensitive to the choice of window size and polynomial order. Inappropriate parameter selection may distort narrow Raman peaks, attenuate weak spectral biomarkers, or alter relative peak intensities.

Wavelet-based denoising methods (Chen et al., 2018) have become popular in Raman spectroscopy due to their ability to separate high-frequency noise from biologically relevant spectral structures across multiple scales. Wavelet denoising decomposes the spectrum into frequency components and suppresses coefficients associated with noise while preserving localized spectral peaks. Compared with polynomial smoothing methods, wavelet approaches can better preserve narrow Raman bands and complex spectral morphologies, particularly in spectra with low signal-to-noise ratios. However, denoising performance depends strongly on the selected wavelet family, decomposition level, and thresholding strategy, which may introduce variability across studies.

Alternative denoising approaches include total variation and sparse regularization methods (Liu et al., 2014), which aim to reduce noise while preserving sharp spectral transitions and peak boundaries. These methods are suitable for spectra containing sparse or narrow spectral features, although their performance depends strongly on regularization parameters and assumptions regarding spectral smoothness. In practice, no single denoising strategy consistently outperforms all others across biomedical Raman datasets, and method selection often is application-dependent. Despite their theoretical advantages, total variation methods remain less commonly used in biomedical Raman workflows than polynomial or wavelet-based denoising approaches.

Recent advances in deep learning have enabled data-driven denoising strategies capable of learning complex spectral priors directly from training data (Lussier et al., 2020). Convolutional and encoder-decoder deep learning architectures (Kazemzadeh et al., 2022a, 2022b; Sjöberg et al., 2025; Zeng et al., 2023) have demonstrated strong performance in recovering Raman peaks from heavily corrupted spectra while reducing processing time through GPU acceleration. Unlike classical filtering approaches, deep learning models can jointly model fluorescence, detector noise, and nonlinear spectral distortions. However, these methods often rely on synthetically generated training spectra, which may inadequately represent the complexity of real biomedical Raman datasets and can therefore limit cross-dataset generalizability. More recently, self-supervised and noise-to-noise learning strategies have gained popularity. These approaches exploit redundancy within spectral datasets to learn denoising transformations directly from noisy observations, potentially improving robustness in settings where clean reference spectra are unavailable.

Denoising procedures may additionally influence spectral interpretability by modifying peak morphology and relative intensity distributions, thereby affecting downstream feature attribution and explainable AI analyses. Importantly, denoising methods should be evaluated not only by signal reconstruction metrics, but also by their impact on downstream tasks such as clustering, classification, and spectral interpretability.

## Baseline Correction

Baseline correction is one of the most critical and challenging preprocessing stages in biomedical Raman spectroscopy because biological samples often exhibit strong fluorescence backgrounds that can exceed Raman signal intensity by several orders of magnitude (Butler et al., 2016). Additional baseline distortions may arise from tissue autofluorescence, instrumental response, sample heating, and acquisition variability (Movasaghi et al., 2007; Pence and Mahadevan-Jansen, 2016). These effects can obscure weak Raman peaks, distort relative peak intensities, and substantially influence downstream statistical and machine learning analyses. Moreover, although fluorescence background is commonly treated as an unwanted artifact, several studies have suggested that fluorescence intensity and morphology may themselves contain biologically relevant information. Aggressive baseline suppression may therefore remove diagnostically informative signals in certain biomedical applications. Baselines are also strongly influenced by excitation wavelengths. Shorter excitation wavelengths often produce stronger Raman scattering but substantially increased fluorescence backgrounds, whereas near-infrared excitation reduces fluorescence at the cost of weaker Raman signal intensity and longer acquisition times (Krafft and Popp, 2015).

Polynomial baseline correction is one of the most widely used approaches in Raman spectroscopy because of its simplicity and computational efficiency. These methods model the slowly varying fluorescence background using low-order polynomial functions that are subsequently subtracted from the measured spectrum. Variants such as ModPoly (Lieber and Mahadevan-Jansen, 2003), IModPoly (Zhao et al., 2007), and airPLS (Zhang et al., 2010) differ primarily in how the baseline is estimated and updated iteratively. Polynomial methods may struggle in the presence of highly complex fluorescence backgrounds and require careful parameter selection to avoid distortion of biologically relevant spectral features.

Morphology-based approaches estimate the baseline through geometric operations that separate broad background components from narrow Raman peaks. Methods such as rolling-ball correction Sternberg, 1983) and BubbleFill (Sheehy et al., 2023) are often effective when fluorescence backgrounds exhibit complex local structure and may outperform simple polynomial models in heterogeneous biological samples. However, their performance remains sensitive to parameter choices and spectral characteristics.

The extended multiplicative signal correction (EMSC) method (Afseth and Kohler, 2012; Martens and Stark, 1991) addresses baseline effects while simultaneously correcting for multiplicative and additive spectral variability. By explicitly modeling sources of spectral variation, EMSC can improve comparability across samples and acquisition conditions, making it particularly attractive for machine-learning applications involving multiple cohorts, instruments, or laboratories.

Baseline characteristics vary substantially depending on tissue type, excitation wavelength, acquisition protocol, and instrumentation. Consequently, baseline correction is highly application-dependent and often requires empirical optimization and careful visual inspection to avoid distortion of biologically relevant spectral information. Different correction strategies may substantially alter spectral intensities, peak relationships, and can substantially affect downstream clustering, classification, and feature attribution results. Consequently, baseline correction should be selected and reported carefully, as seemingly minor methodological differences can influence clustering results, classification accuracy, biomarker identification, and reproducibility across studies.

## Cosmic Ray Removal

Cosmic ray artifacts are high-intensity spikes generated when energetic particles interact with the detector during spectral acquisition. In Raman spectroscopy, these artifacts may substantially exceed genuine Raman peak intensities and introduce artificial spectral features unrelated to the biochemical composition of the sample. Their removal is therefore essential for reliable spectral interpretation, particularly in biomedical datasets where weak Raman biomarkers may already exhibit low signal-to-noise ratios. Detection remains challenging because cosmic rays often resemble narrow Raman peaks and may span one or several adjacent wavelengths.

Most cosmic-ray removal approaches rely on detecting narrow intensity spikes that are inconsistent with the local spectral structure. Detection strategies include neighborhood-based comparisons (Cappel et al., 2010), derivative analysis (Schulze and Turner, 2014), thresholding methods (Whitaker and Hayes, 2018), and statistical outlier detection (Uckert et al., 2019). Once identified, affected points are typically replaced using interpolation or local smoothing procedures. More recently, machine-learning and deep-learning approaches (Kazemzadeh et al., 2022a, 2022b; Sjöberg et al., 2025; Wahl et al., 2020) have been explored for automated artifact detection and correction. These methods can exploit broader spectral context than traditional rule-based algorithms and may improve robustness in large-scale datasets. However, they require carefully curated training data and remain less widely adopted than conventional approaches.

A challenge in cosmic ray removal is avoiding suppression of genuine narrow Raman peaks, particularly in spectra containing sparse or high-frequency spectral structures. Overaggressive artifact removal may therefore eliminate diagnostically relevant biochemical information, especially in surface-enhanced Raman spectroscopy and low-signal biomedical applications.

## Preprocessing Order and Pipeline Design

The order in which preprocessing methods are applied can substantially influence spectral quality and downstream analytical performance (Butler et al., 2016). Preprocessing operations are not independent, and sequential application of denoising, cosmic ray removal, baseline correction, normalization, and scaling may alter both spectral morphology and statistical distributions. Consequently, inappropriate preprocessing order can introduce artifacts, suppress diagnostically relevant biochemical features, or distort downstream machine learning analyses.

Cosmic ray removal is typically performed prior to baseline correction because high-intensity spikes may interfere with estimation of the underlying fluorescence background. Similarly, aggressive smoothing applied before artifact correction may spread localized spectral distortions across adjacent wavelengths, complicating subsequent preprocessing stages. The optimal preprocessing sequence therefore depends strongly on dataset characteristics, acquisition conditions, and the analytical objectives of the study. These challenges are important in large biomedical Raman datasets, where manual inspection of individual spectra is infeasible. As a result, recent machine learning and deep learning approaches increasingly attempt to integrate multiple preprocessing stages within unified end-to-end frameworks, reducing dependence on manually optimized preprocessing pipelines and potentially improving reproducibility across datasets.

## Learning-Based Preprocessing Methods

Machine learning approaches have increasingly been applied to Raman spectral preprocessing in order to automate denoising, baseline correction, and artifact removal within unified computational frameworks (Luo et al., 2022; Lussier et al., 2020). Compared with classical preprocessing pipelines based on manually selected algorithms and hyperparameters, machine learning models can learn complex nonlinear relationships directly from spectral data and may improve scalability in large biomedical datasets. These approaches are effective in Raman spectroscopy because preprocessing stages are often strongly interdependent and highly sensitive to dataset-specific characteristics.

Bayesian learning approaches have been explored for simultaneous denoising and baseline correction. For example, (Li et al., 2022) proposed a sparse Bayesian framework capable of separating Raman peaks from fluorescence and noise components using probabilistic spectral modeling. Their model utilized synthetic spectra, made out of gaussian peaks and noise, and sinusoidal baselines. Deep neural networks have demonstrated strong performance for baseline correction and spectral denoising, through densely connected and convolutional architectures trained on synthetic and experimental Raman datasets (Wahl, 2022; Wahl et al., 2022, 2020). The model presented by (Barton et al., 2021) focused on denoising in Raman spectra, providing a convolutional network for increasing the SNR. Encoder–decoder architectures such as U-Net have further extended deep learning preprocessing by enabling simultaneous denoising and baseline correction within end-to-end frameworks (Kazemzadeh et al., 2022a, 2022b). However, many existing architectures remain constrained by fixed spectral dimensions, reliance on synthetic training data, and limited interpretability regarding the spectral components removed during preprocessing.

More recent architectures have expanded this paradigm by simultaneously decomposing spectra into Raman peaks, baseline components, noise, and cosmic ray artifacts while supporting variable-length spectral inputs (Sjöberg et al., 2025). Such approaches move preprocessing toward unified spectral decomposition frameworks rather than isolated correction steps.

An alternative strategy involves reducing dependence on explicit preprocessing through augmentation-based training. For example, (Liu et al., 2014) proposed exposing neural networks to synthetic baseline distortions and spectral perturbations during training in order to improve robustness to acquisition variability. From a machine learning perspective, this can be seen as enforcing invariance to fluorescence, intensity scaling, and baseline variation. However, excessive augmentation may also encourage models to ignore diagnostically relevant spectral features, and further validation is needed to determine whether augmentation-based robustness generalizes reliably across biomedical datasets.

An unresolved challenge for machine learning preprocessing models is robustness to domain shift arising from differences in instrumentation, excitation wavelength, acquisition protocols, and patient populations (Blake et al., 2022; Ntziouni et al., 2022; Semmelrock et al., 2025). Models trained on highly standardized datasets may exhibit substantially reduced performance when applied across laboratories or clinical environments. The limited interpretability of deep preprocessing models additionally complicates assessment of whether diagnostically relevant biochemical information is preserved during spectral correction.

## Normalization and Scaling for Machine Learning Analysis

Spectral normalization is commonly applied to stabilize optimization and reduce sensitivity to absolute intensity variation. One frequently used approach is L2 normalization, in which each spectrum is divided by its Euclidean norm such that the overall vector magnitude becomes one. This transformation reduces variability arising from differences in acquisition intensity while preserving relative spectral structure.

Another possibility is the z-score standardization, in which intensities at each wavelength are normalized using the mean and standard deviation estimated across the training dataset. This transformation centers spectral features around zero mean and unit variance, preventing high-intensity wavelengths from disproportionately influencing downstream machine learning models.

Although normalization can improve numerical stability and reduce acquisition-related variability, it may also suppress biologically meaningful differences associated with absolute Raman intensity or concentration-dependent effects. Consequently, normalization strategies should be selected carefully depending on whether relative spectral morphology or absolute signal magnitude is expected to carry diagnostic information.

# Unsupervised Structure Discovery in Raman Data

Clustering seeks to identify biologically meaningful organization within spectral data, including tissue compartments, cellular populations, molecular phenotypes, and disease-associated subregions, without requiring extensive manual pathological annotation (Butler et al., 2016). These approaches are useful in oncology because tumor boundaries, infiltrative margins, necrotic regions, and heterogeneous microenvironments often exhibit distinct biochemical and spectral characteristics. After preprocessing has reduced noise and technical variability, clustering methods provide a first layer of biological interpretation by identifying recurring spectral patterns and tissue subpopulations without requiring predefined labels. Clustering can reveal intrinsic biochemical organization and identify previously unrecognized tissue subpopulations or metabolic microenvironments.

Clustering of Raman spectra is challenging due to high spectral dimensionality, fluorescence variability, noise sensitivity, overlapping biochemical signatures, and substantial intra- and inter-tissue heterogeneity (Pence and Mahadevan-Jansen, 2016). In biological tissues, Raman spectra frequently represent mixtures of multiple molecular components rather than discrete biochemical states, resulting in continuous transitions between spectral phenotypes rather than clearly separated clusters. Neighboring Raman wavelengths are often highly correlated, while diagnostically relevant variations may be confined to relatively subtle spectral differences. Consequently, successful Raman clustering frequently requires careful preprocessing, dimensionality reduction, and integration of spatial or manifold-based information in order to obtain biologically meaningful tissue segmentation.

In biomedical Raman spectroscopy, partition-based methods such as K-means, Gaussian Mixture Models, and Fuzzy C-Means remain the most widely used approaches because they provide a practical balance between interpretability, computational efficiency, and performance. Other methods, including hierarchical, graph-based, density-based, and scalable clustering approaches, are typically employed when specific dataset characteristics make them advantageous.

## Partition-Based Clustering Methods

K-means (MacQueen, 1967) is one of the most widely used clustering methods in biomedical Raman spectroscopy due to its simplicity, computational efficiency, and scalability to large hyperspectral imaging datasets. The method partitions spectra into K clusters according to their similarity to cluster centroids and is frequently combined with dimensionality-reduction techniques such as PCA to improve cluster separability in high-dimensional spectral data. K-means assumes that data comes in clusters that are compact and well separated, assumptions that are often violated in complex biological systems. These limitations become particularly evident in oncological applications, where Raman spectra often exhibit substantial overlap between tumor and surrounding tissue.

Consequently, hard assignments may fail to capture mixed tissue states and infiltrative tumor regions.

Representative applications have demonstrated that K-means clustering can successfully delineate biochemical tissue regions in Raman maps, including stromal, necrotic, and tumor-associated areas (Mo et al., 2024). However, performance often deteriorates in highly infiltrative architectures characterized by dispersed malignant cells, weak spectral contrast, or low signal-to-noise ratios, where fine tissue structures may become misclassified. Despite these limitations, K-means and its scalable variants such as Mini-Batch K-means enable clustering of hundreds of thousands of spectra and have been used to characterize spatial intratumoral heterogeneity in glioma Raman imaging (Lita et al., 2024). When combined with dimensionality reduction methods, these approaches can identify subtle infiltration patterns and support downstream supervised analyses.

Gaussian Mixture Models (GMMs) (Reynolds, 2009) are probabilistic clustering methods that assign spectra to clusters with associated membership probabilities rather than enforcing hard cluster boundaries (Hedegaard et al., 2011). Compared with centroid-based approaches such as K-means, GMMs assumes clusters with different covariance structures and are therefore better suited to heterogeneous spectral distributions. This property is relevant in biomedical Raman spectroscopy, where spectra often reflect mixtures of multiple biochemical components rather than discrete tissue classes. GMMs nevertheless present several practical challenges. The number of mixture components must usually be determined in advance or estimated through model-selection criteria such as the Bayesian Information Criterion. In addition, the method is computationally more demanding than K-means and may become unstable in high-dimensional Raman datasets, often requiring dimensionality reduction or regularization strategies.

Fuzzy C-Means (FCM) clustering (Bezdek et al., 1984) is a soft partitioning method that is particularly well suited to biomedical Raman spectroscopy because it allows spectra to belong simultaneously to multiple clusters. Rather than enforcing rigid cluster boundaries, FCM assigns continuous membership values reflecting the degree of association between a spectrum and different tissue populations. This property is especially relevant in oncology, where infiltrative tumors, necrotic regions, edema, and transitional microenvironments frequently produce overlapping biochemical signatures. Consequently, FCM can provide a more realistic representation of tissue heterogeneity than hard clustering approaches such as K-means. FCM is sensitive to noise, initialization, and parameter selection, particularly the fuzzification parameter that controls the degree of cluster overlap. Excessive fuzziness may reduce cluster separability, whereas overly restrictive settings may approach hard-clustering behavior. As a result, robust preprocessing is essential, since noise and baseline artifacts may otherwise become incorporated into cluster memberships.

## Hierarchical and Graph-Based Clustering

Hierarchical and graph-based approaches are particularly valuable when Raman datasets exhibit multiscale or nonlinear organization. Their ability to capture complex biochemical relationships complements the efficiency of partition-based methods, although increased computational requirements may limit their applicability to very large datasets.

Agglomerative clustering (Johnson, 1967) is widely used in biomedical Raman spectroscopy because it can reveal hierarchical relationships between spectra without requiring prior assumptions about the number of tissue classes. This method progressively merges spectrally similar groups, producing a dendrogram that captures multiscale biochemical organization within the data. This is particularly valuable in cancer studies, where the number of tissue classes or tumor subtypes is often unknown and hierarchical relationships may provide insight into intratumoral heterogeneity and biochemical gradients. An additional advantage of agglomerative clustering is its flexibility with respect to similarity measures and linkage strategies, allowing adaptation to the statistical and biochemical characteristics of Raman datasets. Unlike centroid-based approaches, hierarchical clustering does not impose strong assumptions regarding cluster geometry and can therefore accommodate complex spectral structures.

Agglomerative clustering is sensitive to noise, outliers, and early merging decisions, which cannot be reversed once made. These challenges are particularly relevant in Raman spectroscopy, where fluorescence background, instrumental variability, and preprocessing artifacts may affect spectral similarity. Agglomerative clustering is also computationally demanding in terms of both time and memory, limiting its scalability for large Raman datasets. Nevertheless, it remains a valuable exploratory tool because it can reveal intrinsic biochemical substructures and hierarchical relationships that may not be captured by simpler partition-based methods.

Spectral clustering (Ng et al., 2001) is significant for biomedical Raman spectroscopy because it can identify nonlinear and manifold-like cluster structures that are difficult to capture using conventional distance-based methods such as K-means. By exploiting connectivity relationships between spectra rather than relying solely on geometric proximity in the original feature space, spectral clustering can reveal subtle biochemical organization within complex tissue datasets. The method constructs a similarity graph between spectra and uses spectral decomposition to obtain a low-dimensional representation that preserves the intrinsic connectivity structure of the data. Clustering is then performed within this embedding space. The performance of this method is sensitive to the choice of similarity function and associated parameters, which strongly influence the resulting graph structure and cluster composition. In addition, construction and decomposition of large similarity graphs can become computationally demanding, limiting scalability in very large hyperspectral Raman datasets.

## Density-Based Clustering Methods

Density-based clustering methods are well suited in biomedical Raman spectroscopy because they group spectra according to local density structure rather than centroid proximity. Unlike partition-based approaches, these methods can identify complex biochemical organization while naturally accommodating noise and outliers. Such properties are relevant in hyperspectral Raman imaging datasets characterized by heterogeneous tissue organization, infiltrative tumor regions, and substantial spectral variability. However, performance remains sensitive to density-estimation parameters and may degrade in strongly overlapping or sparsely sampled datasets.

DBSCAN (Density-Based Spatial Clustering of Applications with Noise) (Ester et al., 1996) identifies clusters as dense regions separated by sparse areas and explicitly labels outliers as noise. Unlike centroid-based methods, it does not require the number of clusters to be specified in advance and can delineate irregular biochemical tissue structures with complex spatial organization. A major limitation of DBSCAN is its sensitivity to the neighborhood radius and the minimum neighborhood size. Inappropriate parameter choices may fragment biologically related tissue regions or merge distinct biochemical structures. In addition, DBSCAN performs poorly when clusters exhibit substantially different local densities, a common feature of heterogeneous tumor microenvironments.

OPTICS (Ordering Points To Identify the Clustering Structure) (Ankerst et al., 1999) extends DBSCAN by analyzing density structure across multiple scales, making it suitable for datasets containing clusters with varying local densities. This flexibility is valuable in biomedical Raman spectroscopy, where different tissue regions may exhibit substantially different spectral densities due to intratumoral heterogeneity, infiltrative growth patterns, or variable signal quality. Although more flexible than DBSCAN, OPTICS is computationally demanding and can be difficult to interpret. Reliable density estimation may also become challenging in noisy datasets or when spectral populations strongly overlap.

Mean-Shift (Comaniciu and Meer, 2002) is a non-parametric density-based method that identifies clusters by iteratively moving spectra toward regions of maximal local density. It is useful for exploratory analysis when the number of biochemical subpopulations is unknown. In biomedical Raman spectroscopy, Mean-Shift is applicable for exploratory analysis of tissue datasets in which the number of biochemical subpopulations is unknown. Mean-Shift is computationally expensive and highly sensitive to the bandwidth parameter controlling density estimation. Small bandwidths may generate excessive fragmentation, whereas large bandwidths may merge distinct biochemical structures. Performance may further deteriorate in noisy or high-dimensional Raman datasets.

## Scalable Clustering for Large-Scale Raman Datasets

Advances in hyperspectral Raman imaging have substantially increased the size and complexity of biomedical Raman datasets. Modern tissue maps may contain hundreds of thousands or even millions of spectra, creating computational challenges for conventional clustering algorithms. Graph-based and density-based clustering approaches are particularly affected because construction of similarity matrices or neighborhood structures may become computationally prohibitive for whole-slide Raman datasets. Combined preprocessing, dimensionality reduction, and clustering workflows further increase computational and storage requirements. These challenges have motivated the development of approximate clustering strategies, hierarchical summarization methods, and parallelized computational pipelines.

Several strategies have been developed to improve the scalability of Raman clustering workflows. Mini-Batch K-means reduces computational costs by processing subsets of spectra rather than the full dataset, while hierarchical summarization approaches such as BIRCH (Balanced Iterative Reducing and Clustering using Hierarchies) (Zhang et al., 1996) construct compact representations of large spectral collections that can be clustered efficiently. These approaches enable rapid exploratory analysis, coarse tissue segmentation, and preprocessing of large hyperspectral datasets prior to more computationally intensive downstream analyses. However, scalability is often achieved at the cost of simplifying assumptions regarding cluster structure, which may limit performance in Raman datasets characterized by nonlinear spectral organization, overlapping tissue populations, or irregular biochemical manifolds.

Large-scale Raman analysis increasingly relies on GPU acceleration, distributed computing, and parallel processing frameworks. Dimensionality reduction methods such as PCA and UMAP further reduce memory requirements and accelerate downstream clustering by compressing high-dimensional spectral representations.

These developments are transforming Raman clustering from a small-scale exploratory tool into a scalable framework capable of supporting whole-slide tissue analysis and future real-time biomedical applications.

## Validation and Biological Interpretation of Raman Clusters

Validation of clustering results is a major challenge in biomedical Raman spectroscopy because unsupervised methods may identify mathematically separable groups that do not necessarily correspond to biologically meaningful tissue states (Butler et al., 2016; Krafft and Popp, 2015). In hyperspectral Raman imaging, spectral variability may arise from genuine biochemical heterogeneity but also from fluorescence background, preprocessing artifacts, acquisition variability, and instrumental noise. Consequently, interpretation of Raman clusters requires both computational validation and biological corroboration (Hano et al., 2024).

Internal validation metrics such as silhouette scores, Davies–Bouldin indices, cluster compactness, and stability analysis are frequently used to assess clustering quality. However, these measures primarily quantify geometric separation in feature space and may not reflect biological relevance. In Raman datasets, strongly separated clusters may still correspond to continuous biochemical gradients rather than discrete tissue states.

Biological validation therefore plays a central role in Raman tissue analysis. Clustering results are commonly compared with histopathological annotations, immunohistochemistry, molecular profiling, or expert pathological assessment to determine whether identified spectral regions correspond to meaningful tissue structures. Spatial correspondence with tumor boundaries, necrotic regions, stromal compartments, or infiltrative margins provides strong evidence that clustering captures genuine biochemical organization. Increasingly, Raman clustering results are also integrated with multimodal imaging, spatial transcriptomics, or molecular pathology frameworks in order to improve biological interpretation of spectral tissue organization.

Interpretation of Raman clusters also requires analysis of the underlying spectral signatures. Characteristic Raman bands associated with lipids, proteins, nucleic acids, collagen, or metabolic markers may help explain cluster separation. However, spectral interpretation remains challenging because Raman peaks often overlap, molecular assignments are not always unique, and clustering results may depend strongly on preprocessing and dimensionality-reduction strategies (Movasaghi et al., 2007; Pence and Mahadevan-Jansen, 2016).

Reproducibility represents an additional challenge for unsupervised Raman analysis. Clustering results may vary across datasets, instruments, acquisition protocols, preprocessing pipelines, and parameter choices, while biological variability between patients further complicates cross-study comparisons. Robust validation therefore requires assessment across independent cohorts and experimental conditions, particularly in translational applications (Beleites et al., 2013; Blake et al., 2022; Semmelrock et al., 2025).

## Choosing a Clustering Strategy for Raman Data

No single clustering method is optimal for all Raman spectroscopy applications. Method selection depends on dataset size, spectral complexity, expected cluster structure, and the biological question being investigated. The major clustering paradigms used in biomedical Raman spectroscopy differ substantially in their assumptions, scalability, and suitability for different biological objectives. Table 2 summarizes their typical application domains and practical trade-offs.

Partition-based methods such as K-means remain attractive for large-scale Raman datasets because of their simplicity, computational efficiency, and ease of interpretation. Variants such as Mini-Batch K-means further improve scalability for large hyperspectral imaging datasets. In contrast, soft-clustering approaches such as Gaussian Mixture Models and Fuzzy C-Means are often better suited to gradual biological transitions and heterogeneous tissue organization.

Density-based and graph-based approaches provide additional flexibility when cluster boundaries are complex or unknown. Methods such as DBSCAN and OPTICS can identify irregular cluster shapes and detect outliers without requiring the number of clusters to be specified in advance, making them useful for exploratory analyses and heterogeneous tissue samples. Spectral clustering is particularly effective when biologically meaningful groups are separated by nonlinear relationships that cannot be adequately represented by simple distance-based methods, although its computational requirements may limit its use for very large datasets.

Ultimately, clustering results should be evaluated not only using computational metrics but also in terms of their biological relevance. The most informative clusters are those that correspond to meaningful tissue structures, cellular populations, molecular phenotypes, or pathological regions. Consequently, successful Raman clustering studies often combine unsupervised learning with complementary sources of information such as histopathology, molecular profiling, spatial localization, or expert annotation in order to establish biological validity and clinical significance.

*Table 2. Comparison of clustering methods commonly used in biomedical Raman spectroscopy. Method selection depends on dataset size, expected cluster structure, biological objectives, and computational constraints.*

| Raman analysis objective | Recommended methods | Main advantages | Main limitations |
|---|---|---|---|
| General purpose tissue segmentation | K-means | Simple, fast, easy to interpret | Predefined number of clusters; assumes simple cluster structure |

| Large Raman maps | Mini-Batch K-means, BIRCH | Fast, scalable | Assume simple cluster structure |
|---|---|---|---|
| Tumor heterogeneity | FCM, GMM | Soft assignments | Higher computational cost |
| Unknown number of populations | DBSCAN, OPTICS | Predefined number of clusters not needed | Parameter sensitive |
| Complex nonlinear structure | Spectral clustering | Captures manifold structure | Scalability limitations |
| Exploratory analysis | Hierarchical clustering | Easy visualization | Computationally expensive |

# Supervised Learning for Diagnosis and Molecular Stratification

Classification represents one of the most successful applications of machine learning in biomedical Raman spectroscopy. It aims to associate Raman measurements with predefined biological, molecular, or clinical outcomes. Current applications span a wide range of objectives, including disease diagnosis, molecular characterization, treatment-relevant stratification, and real-time clinical decision support (Pence and Mahadevan-Jansen, 2016). The following subsections examine these application domains and illustrate how machine learning is extending the role of Raman spectroscopy from diagnostic discrimination toward increasingly informative and clinically actionable decision-support systems.

## Disease Detection and Diagnostic Classification

Raman spectroscopy combined with machine learning enables non-invasive diagnosis from a wide range of biological samples, including cerebrospinal fluid, saliva, serum, and plasma. Among the various biomedical applications of Raman spectroscopy, liquid biopsy has emerged as one of the most active areas of machine learning–assisted classification. By capturing subtle molecular alterations associated with disease, Raman-based models have demonstrated strong diagnostic performance across neurological, oncological, and cardiovascular disorders. In addition to conventional biofluids, recent studies have explored less common sample types such as wet plasma and cerumen, further expanding the range of potential clinical applications.

Representative applications are quite diverse. In neurodegenerative disorders, Raman spectroscopy combined with machine learning has been used to classify Alzheimer's disease from both cerebrospinal fluid and saliva samples (Ralbovsky et al., 2019). In oncology, plasma- and serum-based approaches have demonstrated strong performance for ovarian, lung, and thyroid cancer detection, using models ranging from conventional chemometric methods such as PCA-LDA and PLS-LDA to artificial neural networks (Chen et al., 2022; Zhang et al., 2024; Xia et al., 2021). Plasmon-enhanced Raman spectroscopy coupled with machine learning has also been explored for rapid cardiovascular risk assessment and early detection of acute myocardial infarction (Liu et al., 2023). Recent studies further expanded Raman-based diagnostics to less conventional biofluids, including wet plasma and cerumen, highlighting the versatility of the technology while emphasizing the need for standardized acquisition and preprocessing protocols (Farnesi et al., 2025).

Direct tissue analysis is one of the most mature applications of Raman spectroscopy in medicine. Tissue-based measurements preserve spatial and microenvironmental information, enabling machine learning models to exploit subtle biochemical differences between healthy and diseased regions. By exploiting disease-specific biochemical alterations within the tissue microenvironment, Raman-based models can discriminate malignant and non-malignant tissues while supporting subtype classification. Applications in breast and lung cancer demonstrate clinically relevant diagnostic performance using both conventional machine-learning and deep-learning approaches (Zhang et al., 2022; Yan et al., 2024). Benchmarking and meta-analytic studies further support the robustness of Raman-based cancer classification, although substantial variability remains across datasets, acquisition protocols, and validation strategies (Sharma et al., 2025). Across multiple tumor types, including breast, lung, colorectal, liver, and salivary gland cancers, both deep-learning and chemometric methods have demonstrated strong diagnostic performance (Czaplicka et al., 2021). Recent studies have also explored serum-based Raman diagnostics for liver cancer (T. Sun et al., 2024). Using surface-enhanced Raman spectroscopy (SERS) to detect low-abundance serum proteins, combined with PCA-LDA and PLS-SVM classifiers, high diagnostic accuracy was achieved across multiple disease stages. These results further highlight the potential of minimally invasive Raman-based liquid biopsy approaches for early cancer detection and screening.

Raman spectroscopy has emerged as a promising tool also for microbiological applications, where rapid and accurate pathogen identification remains a major clinical challenge. Machine learning models can classify bacterial species directly from their spectral fingerprints, avoiding the lengthy culture-based procedures commonly used in clinical microbiology. Recent advances in SERS have further improved sensitivity, enabling strain-level discrimination and antimicrobial susceptibility testing from complex biological samples. Studies by (Chen et al., 2025) demonstrate that Raman-based microbiology can provide rapid, culture-independent identification of pathogens while simultaneously capturing metabolically relevant information associated with antimicrobial resistance and bacterial physiology. Raman spectroscopy combined with machine learning has also been applied to host-response characterization. For example, serum Raman spectra have been used to distinguish patients with chronic sinusitis from healthy controls (Raczkiewicz et al., 2026), demonstrating that disease-associated systemic biochemical alterations can be detected through blood-based Raman analysis. Such studies illustrate that Raman-based diagnostics are increasingly extending beyond direct pathogen detection toward characterization of inflammatory and immune-related disease states.

## Molecular Characterization and Precision Medicine

A major trend in contemporary Raman research is the transition from disease detection toward molecular characterization. Rather than simply distinguishing diseased from healthy tissue, machine-learning models increasingly aim to identify molecular subtypes, treatment-relevant biomarkers, and therapy-resistant phenotypes. This shift is particularly evident in oncology, where molecular classification increasingly guides prognosis and treatment selection. Raman spectroscopy has been applied to subtype classification across multiple cancer types, with breast cancer providing one of the most extensively studied examples. In breast cancer, Raman spectroscopy combined with PCA-SVM, PCA-DFA, and deep-learning approaches has been used to distinguish major molecular subtypes, including luminal A, luminal B, HER2-positive, basal, and triple-negative tumors (Li et al., 2024).

Similar developments are evident in neuro-oncology, where Raman spectroscopy has been applied to molecular classification of gliomas, including clinically relevant markers such as IDH mutation status and methylation-defined subtypes. Recent studies indicate that Raman-derived molecular signatures can support clinically meaningful stratification and rapid label-free molecular characterization of diffuse gliomas (Hollon et al., 2023). Molecularly oriented Raman classification has also been explored in gastric, thyroid, and skin cancers, although cross-study heterogeneity and limited external validation remain persistent barriers to clinical implementation (Noh et al., 2025; Bellantuono et al., 2023; Zhao et al., 2024).

Raman spectroscopy increasingly supports characterization of treatment response and disease heterogeneity, as shown by ovarian cancer studies distinguishing cisplatin-resistant phenotypes and by exosome-based Raman analysis for minimally invasive molecular profiling (Y.-J. Li et al., 2026; Parlatan et al., 2023).

Hematological malignancies provide another important application area for molecular Raman analysis because measurements can capture subtle molecular signatures from biofluids and individual cells. Deep learning–assisted SERS platforms have enabled rapid classification of acute leukemia subtypes from low-volume cerebrospinal fluid samples (D. Zhang et al., 2025), while chemometric approaches have been used to distinguish molecular subtypes of acute lymphoblastic leukemia associated with specific genetic alterations (Adamczyk et al., 2024). These studies further illustrate the potential of Raman spectroscopy to support precision medicine through molecularly informed disease stratification.

## Real-Time and Intraoperative Decision Support

Raman spectroscopy is promising in intraoperative settings, where rapid, label-free tissue characterization can directly influence clinical decision-making. Advances in instrumentation, imaging, and machine learning have substantially reduced analysis times while maintaining high diagnostic accuracy, bringing Raman-based decision support closer to routine clinical use. In neurosurgery, Raman systems have demonstrated accurate classification of brain tumors and near real-time detection of tumor infiltration (Kondepudi et al., 2025). Similar results have been reported in meningioma, pediatric brain tumors, and soft-tissue sarcomas (Dulude et al., 2025), suggesting broad applicability across surgical oncology.

One of the most influential demonstrations of clinical translation is the work of (Hollon et al., 2023), who combined rapid label-free optical imaging with artificial intelligence to perform molecular classification of diffuse gliomas during surgery. By delivering molecularly informative classifications within operative time constraints, this study established an important benchmark for real-time Raman-guided neurosurgery and clinical deployment.

The translation of Raman spectroscopy toward real-time clinical decision support is not limited to neuro-oncology. In colorectal disease, automated machine-learning pipelines combined with portable Raman instrumentation have demonstrated the feasibility of in vivo tissue classification during clinical examination. For example, (Vališ et al., 2024) evaluated an automated Raman-based classification framework in 377 subjects and demonstrated accurate real-time discrimination of colorectal pathology. Together, these studies suggest that Raman-guided decision support may become feasible across a broad range of clinical settings.

Spatially offset Raman spectroscopy (SORS) provides another example of how advances in Raman acquisition can support clinical translation (Mosca et al., 2021; K. Zhang et al., 2025). By probing subsurface tissue layers and reducing the influence of superficial tissue components, SORS enables non-invasive assessment of deeper pathological structures. Recent studies have demonstrated improved discrimination of non-melanoma skin cancers using SORS compared with conventional back-scattering Raman measurements, highlighting the potential of advanced Raman modalities for future point-of-care and in vivo diagnostic applications.

Continued progress in instrumentation and machine learning is further expanding the clinical potential of Raman spectroscopy. Wide-field Raman imaging reduces spatial sampling bias by enabling more comprehensive characterization of heterogeneous tissues, while probabilistic machine-learning frameworks provide uncertainty estimates alongside diagnostic predictions, an important requirement for clinical decision support (Kothari et al., 2021).

## Emerging Directions: Representation and Transfer Learning

Most machine-learning approaches in Raman spectroscopy remain strongly supervised and depend on relatively small labeled datasets. However, obtaining large, well-annotated Raman cohorts is expensive and often requires extensive pathological or molecular characterization (Blake et al., 2022; Boateng, 2025). These limitations have motivated increasing interest in representation-learning approaches that aim to extract informative spectral features from large collections of unlabeled or weakly labeled data before downstream classification tasks are performed.

Self-supervised (Jing and Tian, 2021) and contrastive learning (Chen et al., 2020) approaches provide an alternative to conventional supervised training by learning informative spectral representations from unlabeled data. Rather than directly optimizing a classification objective, these methods aim to capture the underlying structure of spectral measurements in a latent representation that can subsequently be adapted to downstream tasks such as diagnosis, subtype classification, or treatment-response prediction. This paradigm is particularly attractive for biomedical Raman spectroscopy because large collections of spectra can often be acquired more easily than high-quality clinical or molecular annotations. By exploiting unlabeled datasets during pretraining, self-supervised approaches may improve sample efficiency, reduce dependence on costly expert labeling, and enhance robustness to biological and technical variability. Recent studies suggest that representation learning can capture spectral features that generalize across related tasks more effectively than models trained solely on a single supervised objective, although systematic evaluation in biomedical Raman spectroscopy remains limited.

Transfer learning has emerged as a promising strategy for addressing one of the central limitations of biomedical Raman spectroscopy (Kamran et al., 2025; Zhang et al., 2020) : the scarcity of large, well-annotated datasets. Rather than training models from scratch for each application, transfer-learning approaches reuse representations learned from larger spectral collections and adapt them to new classification tasks with limited labeled data. Early studies demonstrated that pretrained Raman models improve classification accuracy, convergence, and feature extraction performance when only small target datasets are available. More recently, transfer learning has been increasingly applied to calibration transfer and cross-instrument generalization, enabling models trained on one spectrometer or laboratory setting to be adapted to another (Lai et al., 2025; Q. Li et al., 2026). Closely related domain-adaptation methods (Liu et al., 2024; Z. Zhang et al., 2025) seek to reduce distributional differences arising from instrument variability, batch effects, acquisition protocols, and population heterogeneity. Recent Raman studies have employed unsupervised and multisource domain-adaptation strategies to transfer knowledge across datasets and disease domains while requiring little or no target-domain annotation (Liu et al., 2024; Zhang et al., 2025). These approaches are becoming increasingly important for clinical translation because they directly address one of the major challenges of Raman machine learning: transferring predictive models across laboratories, instruments, and real-world deployment environments.

More recently, foundation-model concepts (Bhatia et al., 2025) have begun to influence spectroscopic machine learning. Although still at an early stage, pretrained spectral encoders and multimodal representation-learning frameworks offer the possibility of learning general-purpose embeddings from large heterogeneous datasets, potentially integrating Raman spectra with microscopy, pathology, molecular profiling, and clinical metadata.

Despite their promise, these approaches remain substantially less mature than conventional supervised learning in biomedical Raman spectroscopy. Their success will depend on the availability of larger datasets, standardized benchmarks, and rigorous evaluation across instruments and clinical settings. Nevertheless, representation learning offers a promising route toward addressing several of the central challenges facing Raman AI, including limited annotation, domain shift, and multimodal data integration.

# Explainability and Raman Biomarker Discovery

High predictive performance alone is rarely sufficient in biomedical applications. Understanding which spectral features drive model predictions is essential for biological interpretation, biomarker discovery, and clinical trust. Unlike many machine-learning domains where input features have limited physical interpretation, Raman spectra directly reflect molecular composition through vibrational signatures associated with proteins, lipids, nucleic acids, and metabolites. Explainability therefore serves a dual purpose in biomedical Raman spectroscopy: increasing confidence in model predictions and facilitating the discovery of disease-associated spectral biomarkers (Pence and Mahadevan-Jansen, 2016).

Recent studies increasingly employ feature-attribution methods to identify spectral regions contributing most strongly to model predictions. Depending on the model architecture, these approaches include feature ranking, saliency maps, Grad-CAM visualizations, attention mechanisms, SHAP values, and related attribution techniques. Recent deep-learning studies have combined convolutional neural networks with Grad-CAM-based visualization to identify Raman shifts driving breast cancer subtype classification (Li et al., 2024), while attention-based architectures have been used to highlight diagnostically informative spectral regions in complex biomedical datasets.

Explainability has increasingly been incorporated directly into model design. For example, interpretable SERS frameworks have been developed for microbial identification (Chen et al., 2025), linking classification decisions to metabolite-level signatures rather than abstract latent representations. Similar efforts increasingly seek to connect pathogen classification and antimicrobial resistance prediction to biologically meaningful molecular signatures.

Explainability is particularly valuable in Raman spectroscopy because, unlike many imaging modalities, the identified predictive features correspond directly to physically measurable spectral bands, potentially connected to biomarker discovery. Rather than identifying only whether a sample belongs to a particular disease class, Raman-based machine-learning models can reveal spectral regions associated with underlying molecular differences. (Huang et al., 2023) linked discriminative Raman peaks identified by deep-learning models to specific biochemical components and spatial tissue organization in liver cancer. Similarly, molecular classification studies (Lita et al., 2024) of diffuse gliomas have identified discriminative Raman wavelengths associated with clinically relevant molecular subtypes, including IDH mutation status, enabling interpretation of predictive models in terms of tumor biochemistry and molecular pathology. Such studies demonstrate how machine learning can guide the identification of candidate Raman biomarkers that may warrant further biochemical investigation and experimental validation.

The relationship between explainability and validation is particularly important in biomedical applications. Candidate biomarkers identified through machine-learning models are rarely accepted solely on the basis of predictive importance. Instead, spectral assignments are typically evaluated against existing biochemical knowledge, pathological findings, molecular profiling data, or complementary analytical techniques. This integration of computational and experimental evidence is essential for distinguishing biologically meaningful biomarkers from dataset-specific artifacts or spurious correlations.

Interpretation of Raman-based machine-learning models nevertheless remains challenging. Neighboring Raman wavelengths often contain highly correlated information, while individual spectral bands may receive contributions from multiple biochemical compounds. Consequently, feature-attribution methods may identify discriminative spectral regions without uniquely determining the underlying molecular mechanism. Attribution results may further depend on preprocessing choices, model architecture, and training variability, complicating comparisons across studies. These limitations highlight the importance of combining computational explanations with independent biochemical, pathological, or molecular validation.

# Multimodal Integration of Raman Spectroscopy

Many clinically relevant questions require integration of molecular, morphological, and clinical information that cannot be captured by Raman spectroscopy alone (Das et al., 2017; Fitzgerald et al., 2023). Consequently, increasing attention has been directed toward multimodal approaches that combine Raman measurements with complementary data sources such as medical imaging, histopathology, molecular profiling, and clinical metadata. By integrating these orthogonal sources of information, multimodal frameworks can improve diagnostic performance, enhance biological interpretation, and support more comprehensive characterization of disease processes. Recent advances in multimodal machine learning (Doan et al., 2026) have further accelerated this trend by enabling joint analysis of spectral, imaging, molecular, and clinical data across multiple biological scales. The growing diversity of multimodal Raman frameworks has led to several distinct categories of integration, each addressing different limitations of standalone Raman spectroscopy; representative examples are summarized in Table 3.

**Table 3**. Representative multimodal Raman spectroscopy frameworks in biomedicine. Multimodal approaches combine Raman spectroscopy with complementary imaging, spectroscopic, pathological, and molecular profiling technologies to overcome limitations of individual modalities. These integrations provide anatomical context, morphological information, enhanced molecular characterization, single-cell resolution, and mechanistic biological interpretation, thereby supporting applications ranging from cancer diagnosis and surgical guidance to biomarker discovery and precision medicine.

| Category | Representative Modalities | Primary Contribution | Example Applications |
|---|---|---|---|
| Structural and anatomical imaging | OCT, MRI, photoacoustic imaging | Provides spatial, morphological, and anatomical context absent from Raman spectra alone | Tumor localization, surgical guidance, tissue characterization |
| Optical microscopy and pathology | Fluorescence microscopy, digital pathology, stimulated Raman histology | Combines biochemical information with cellular and tissue morphology | Cancer diagnosis, histopathology, intraoperative decision support |
| Complementary spectroscopic modalities | FTIR spectroscopy | Expands molecular coverage through complementary vibrational information | Disease classification, biomarker discovery, molecular phenotyping |

| Single-cell and cellular profiling | Raman-activated cell sorting (RACS), Raman2RNA | Characterizes cellular states and heterogeneity at single-cell resolution | Cell-state identification, differentiation studies, microbial phenotyping |
|---|---|---|---|
| Molecular profiling and omics integration | Transcriptomics, spatial transcriptomics, multi-omics platforms, RamanOmics | Links spectral signatures to molecular mechanisms and biological pathways | Precision medicine, biomarker discovery, systems biology |

## Raman and Medical Imaging

Raman spectroscopy provides highly specific biochemical information but limited anatomical and spatial context (Das et al., 2017; Fitzgerald et al., 2023). Complementary imaging modalities, including MRI, OCT, photoacoustic imaging, fluorescence microscopy, and digital pathology, provide structural and functional information that is difficult to infer from spectral measurements alone (Das et al., 2017; Movasaghi et al., 2007). Their integration enables simultaneous characterization of tissue architecture and molecular composition, improving both diagnostic performance and biological interpretability. Consequently, multimodal Raman imaging has become an active area of research in applications ranging from cancer diagnosis and surgical guidance to neurological and cardiovascular disorders (Hollon et al., 2023).

Integration with complementary spectroscopic modalities represents another important direction in multimodal Raman analysis. Among these, the combination of Raman and Fourier-transform infrared (FTIR) spectroscopy is particularly well established. Because Raman and FTIR spectroscopy are governed by complementary selection rules, they probe different aspects of molecular structure and together provide broader biochemical coverage than either modality alone. Recent studies have combined Raman and FTIR spectra within machine-learning frameworks to improve disease classification, biomarker discovery, and molecular phenotyping. For example, multimodal deep-learning architectures have been proposed to jointly analyze Raman and FTIR spectra, demonstrating improved performance in multi-cancer classification compared with single-modality approaches (Krafft and Popp, 2015; Zhou et al., 2024).

Several multimodal imaging paradigms have emerged as particularly influential. Raman–MRI combines molecular characterization with whole-organ anatomical visualization and has been explored extensively in neuro-oncology (Hollon et al., 2023; Jermyn et al., 2016). Raman–OCT integrates biochemical and microstructural information (Fitzgerald et al., 2023), while Raman–fluorescence microscopy (Becker et al., 2021) and Raman–optoacoustic imaging (Chen et al., 2024) provide complementary cellular and spatial information that is difficult to obtain from Raman spectroscopy alone. Together, these approaches illustrate how Raman spectroscopy increasingly functions as one component of a broader multimodal diagnostic framework rather than a standalone technology.

## Raman Imaging and Digital Pathology

The convergence of Raman spectroscopy and digital pathology represents one of the most active areas of multimodal biomedical analysis (Desroches et al., 2018; Hollon et al., 2023). Traditional histopathology remains the clinical gold standard for tissue diagnosis, providing detailed morphological information through microscopic examination of stained tissue sections (Madabhushi and Lee, 2016). However, conventional workflows typically require tissue processing, staining, and expert interpretation, limiting their suitability for rapid intraoperative decision-making. Raman spectroscopy complements histopathology by providing label-free biochemical information directly from tissue samples, enabling integrated analysis of tissue morphology and molecular composition (Lu et al., 2016). Consequently, increasing efforts have focused on combining Raman-derived information with digital pathology workflows to improve tissue characterization, accelerate diagnosis, and support precision medicine applications (Desroches et al., 2018; Hollon et al., 2023; Reinecke et al., 2024).

Recent advances in Raman imaging have enabled the emergence of pathology-oriented workflows that combine molecular specificity with histological visualization (Desroches et al., 2018). Stimulated Raman histology (SRH) generates label-free images that closely resemble conventional hematoxylin and eosin (H&E) sections while avoiding tissue fixation and staining procedures (Lu et al., 2016). When combined with machine-learning and deep-learning models, these approaches enable automated tissue classification, tumor detection, and intraoperative decision support (Daoust et al., 2025). In neuro-oncology, AI-assisted SRH systems have demonstrated the ability to classify brain tumors directly from fresh tissue specimens within clinically relevant time frames, illustrating the potential of Raman-based digital pathology for rapid surgical diagnosis. More recently, generative learning approaches have been used to produce virtual H&E images from Raman data, further narrowing the gap between molecular imaging and conventional pathology workflows (Reinecke et al., 2024).

The clinical potential of Raman-based digital pathology has been demonstrated most prominently in neuro-oncology. Using stimulated Raman histology combined with deep learning, (Hollon et al., 2023) developed a framework capable of performing rapid molecular classification of diffuse gliomas directly from fresh surgical specimens. The system provided diagnostically relevant molecular information within the intraoperative time window, demonstrating that Raman-derived images can support both histopathological assessment and molecular stratification without requiring conventional laboratory processing. More broadly, these studies illustrate how the integration of Raman imaging, computational pathology, and machine learning can transform tissue diagnosis from a sequential workflow into a unified analytical framework that simultaneously captures morphological and molecular characteristics of disease.

Several challenges remain before Raman-based digital pathology can be routinely integrated into clinical workflows. Large-scale validation across institutions is still limited, and differences in instrumentation, acquisition protocols, and tissue preparation may affect model generalizability (Blake et al., 2022; Semmelrock et al., 2025). Furthermore, integration with existing pathology infrastructure requires robust standardization, interoperability, and regulatory validation (Bera et al., 2019). Nevertheless, the ability to combine morphological assessment, molecular characterization, and machine-learning analysis within a single framework positions Raman-based digital pathology as a promising component of future precision diagnostics (Hollon et al., 2023; Lu et al., 2016; Reinecke et al., 2024).

## Raman and Molecular Profiling

Beyond integration with imaging and pathology, Raman spectroscopy is increasingly being combined with molecular profiling technologies, including genomics, transcriptomics, proteomics, metabolomics, and molecular pathology (Butler et al., 2016). Whereas Raman spectra provide a biochemical fingerprint of tissue composition, molecular profiling offers direct information about genetic alterations, signaling pathways, and cellular states (Hasin et al., 2017). The integration of these complementary data sources enables more comprehensive characterization of disease mechanisms and supports the development of precision-medicine approaches (Karczewski and Snyder, 2018). Such multimodal frameworks may also help relate Raman-derived biomarkers to their underlying molecular basis, improving biological interpretation and clinical utility (Hollon et al., 2023; Melitto et al., 2022).

Recent advances (Kobayashi-Kirschvink et al., 2024) suggest that Raman spectroscopy may eventually contribute more directly to molecular-state inference. The Raman2RNA framework introduced by Kobayashi-Kirschvink et al. combines Raman microscopy and deep learning to predict single-cell transcriptomic profiles from label-free Raman measurements. Trained on paired Raman and single-cell RNA-sequencing datasets, the model demonstrated that Raman spectra contain sufficient information to recover aspects of cellular transcriptional state, highlighting the potential of Raman spectroscopy as a non-destructive surrogate for molecular profiling in selected applications. While such approaches remain at an early stage of development, they illustrate a possible future convergence between vibrational spectroscopy, machine learning, and single-cell omics.

Neuro-oncology provides some of the clearest examples of this paradigm. Raman spectroscopy combined with machine learning has been used to identify clinically important molecular characteristics of gliomas, including IDH mutation status and methylation-defined subtypes (Hollon et al., 2023). These studies suggest that molecular alterations generate reproducible biochemical signatures that can be detected directly from Raman spectra. Similar observations have been reported in breast cancer, where Raman-based models distinguish major molecular subtypes including luminal A, luminal B, HER2-positive, basal, and triple-negative tumors (Li et al., 2024). Together, these studies suggest that Raman spectra capture biologically meaningful information extending beyond conventional tissue classification and reflecting underlying molecular organization.

Multimodal integration increasingly supports characterization of disease heterogeneity and treatment response. In ovarian cancer, Raman-based machine-learning models have been used to distinguish cisplatin-resistant and treatment-sensitive phenotypes, demonstrating the potential of spectral analysis to capture biologically relevant differences associated with therapeutic outcomes (Y.-J. Li et al., 2026). Raman spectroscopy has also been applied to the analysis of extracellular vesicles and exosomes, enabling minimally invasive molecular profiling through liquid-biopsy samples (Parlatan et al., 2023). These applications illustrate a broader shift from diagnosis toward functional characterization of disease states, where spectral measurements are interpreted alongside molecular and cellular processes relevant to prognosis and treatment selection.

Recent work has explored the integration of Raman imaging with spatially resolved molecular profiling technologies. The RamanOmics framework (K. Zhang et al., 2025) combines Raman imaging, single-nucleus RNA sequencing, and spatial transcriptomics to construct multimodal tissue atlases that relate biochemical signatures to spatial gene-expression programs. Such approaches aim to bridge molecular composition, tissue architecture, and transcriptional activity within a unified analytical framework, providing new opportunities for studying tissue heterogeneity, ageing, tumor microenvironments, and disease progression. Although still emerging, these studies illustrate how Raman spectroscopy may contribute to next-generation spatial multi-omics analyses.

Integration of Raman spectroscopy with molecular profiling technologies, including genomic, transcriptomic, proteomic, and metabolomic data, represents an important emerging direction for precision medicine (Hasin et al., 2017; Karczewski and Snyder, 2018). Although differences in scale, dimensionality, and data availability present significant challenges, such approaches may help connect Raman-derived phenotypes with underlying molecular mechanisms and improve biological interpretation of spectral biomarkers. Ultimately, multimodal frameworks may bridge rapid phenotypic characterization and molecular profiling, enabling more comprehensive and clinically informative models of disease (Subramanian et al., 2020).

# Towards Robust and Deployable Raman AI

The rapid growth of machine-learning methods for biomedical Raman spectroscopy has produced impressive advances in classification accuracy, molecular characterization, explainability, and multimodal analysis. However, methodological innovation alone is insufficient for clinical translation. Reproducibility, standardized evaluation, software availability, and deployment-oriented workflows are increasingly recognized as critical requirements for transforming promising research prototypes into robust and clinically usable systems. This section discusses the software infrastructure, reproducibility challenges, and community practices needed to support reliable and deployable Raman AI.

## Software and Data Infrastructure

### Open Software Resources

A prerequisite for reproducible machine-learning research in biomedical Raman spectroscopy is the availability of well-documented, openly maintained software that implements common analytical steps in a standardized and interoperable way. Historically, however, Raman analysis has relied heavily on in-house scripts, proprietary instrument software, or loosely documented MATLAB and Python code that is rarely shared alongside publications (Semmelrock et al., 2025). As a result, textual descriptions of methods are often insufficient to reproduce complete preprocessing and modelling workflows. The emergence of dedicated open-source Raman packages, together with the broader scientific Python ecosystem for machine learning, has begun to address this gap, although adoption remains uneven across the literature.

RamanSPy (Georgiev et al., 2024) is the most comprehensive open-source framework developed specifically for Raman spectroscopy. By providing standardized workflows for preprocessing, analysis, visualization, and machine learning within a unified Python environment, it reduces reliance on custom laboratory-specific code and promotes reproducible analysis across studies. Its open-source design, documentation, and example datasets make it an important step toward community-shared analytical infrastructure for biomedical Raman research. Other open-source tools relevant to Raman analysis include SpectroChemPy (Travert and Fernandez, 2025), a multi-modal spectroscopy framework, and HyperTools (Heusser et al., 2018), a visualization package for exploring high-dimensional spectral embeddings. While neither is specifically designed for biomedical Raman machine learning, both can support reproducible data exploration and multimodal analysis workflows.

Several barriers continue to limit reproducibility and deployment in biomedical Raman machine learning. Many studies still do not release analysis code, trained models, or complete computational environments, making independent verification difficult even when methods are described in detail (Semmelrock et al., 2025). In addition, differences in preprocessing implementations, parameter choices, and validation protocols often hinder meaningful comparison between studies. The field also lacks widely accepted benchmark datasets paired with reference implementations, limiting objective evaluation of new methods. Addressing these challenges will require broader adoption of version-controlled code repositories, containerized computational environments, standardized reporting practices, and community benchmarks that support reproducible comparison across laboratories, instruments, and clinical settings (Georgiev et al., 2024).

## File Formats

Raman spectroscopy data are distributed across a fragmented ecosystem of largely proprietary, instrument-specific file formats. Although these formats typically store both spectral data and acquisition metadata, they are often not interoperable across manufacturers and software platforms (Ntziouni et al., 2022). For machine-learning workflows, which increasingly combine data from multiple instruments and studies, this fragmentation creates practical barriers to data sharing, integration, and reproducibility. Data conversion is frequently required before analysis and may result in loss of metadata or undocumented processing steps that complicate independent validation.

Open standards provide a potential solution to this interoperability problem. The JCAMP-DX standard (Grasselli, 1991) supports storage of Raman spectra together with structured acquisition metadata and is supported by several open-source analysis tools, including RamanSPy and SpectroChemPy. Despite these advantages, adoption in biomedical Raman spectroscopy remains limited, with many studies releasing data as metadata-poor CSV files or as proprietary vendor formats. Broader adoption of JCAMP-DX or comparable open standards would improve interoperability, facilitate data sharing, and support large-scale aggregation and meta-analysis of Raman datasets.

Raman imaging datasets introduce additional standardization challenges because each acquisition consists of a hyperspectral data cube rather than a single spectrum. No widely adopted open standard currently exists for storing such data, and published imaging datasets are often distributed either as proprietary binary files or as exported images that do not preserve the full spectral information. The development of an open hyperspectral format analogous to OME-TIFF in fluorescence microscopy (Goldberg et al., 2005) would be a significant step toward enabling reproducible machine learning analysis of Raman imaging data.

## Code Availability

Reproducibility in machine-learning research depends not only on detailed methodological descriptions but also on the availability of code, trained models, and preprocessing pipelines. In the biomedical Raman literature, these resources are still released inconsistently (Semmelrock et al., 2025). As a result, many published workflows cannot be independently verified or directly compared, even when the underlying methods are described in reasonable detail. Small implementation choices, including preprocessing order, parameter settings, and data-handling procedures, can substantially influence model performance, yet often remain inaccessible to readers in the absence of executable code.

A recent systematic benchmarking study of deep learning architectures for Raman spectral classification found that, for a substantial fraction of published models, no official implementation was publicly available and the paper lacked sufficient architectural detail for independent reproduction (Sineesh and Kamsali, 2026). The consequence is that results reported across different groups are difficult to compare, successful approaches propagate slowly, and the community lacks the shared infrastructure needed for incremental progress. Encouraging or requiring code deposition as a condition of publication, together with version-controlled repositories and containerized computational environments, would substantially improve this situation (Semmelrock et al., 2025).

# Standardization and Reproducibility

The long-term clinical utility of machine-learning models for Raman spectroscopy depends not only on predictive performance within a single study, but also on whether results obtained in one laboratory, with one instrument, and on one patient cohort can be trusted, compared, and reproduced by others. Achieving this goal requires several foundational practices that are still inconsistently implemented across the biomedical Raman literature: standardized file formats, explicit acquisition metadata, agreed measurement protocols, and transparent reporting of preprocessing procedures. These dimensions are closely interconnected, and deficiencies in any one of them can undermine reproducibility. Progress across all four fronts is therefore required for the development of clinically deployable and independently validated machine-learning systems (Semmelrock et al., 2025).

### Acquisition Metadata

Reproducibility in Raman spectroscopy begins at the point of data collection. A spectrum is not a self-describing object: its interpretation and comparability depend critically on the conditions under which it was acquired. Measurements obtained using different excitation wavelengths, laser powers, integration times, optical configurations, or sample preparation protocols may differ substantially even when derived from the same tissue type (Workman, 2018). Without complete acquisition metadata, these differences remain invisible to readers and cannot be distinguished from genuine biological variation by downstream machine-learning models.

Both instrumental and biological metadata are essential for reproducible Raman machine-learning studies. Instrumental metadata include information describing acquisition conditions and spectrometer configuration, while biological metadata encompass sample origin, preparation procedures, clinical characteristics, and the assignment of ground-truth labels (Blake et al., 2022). Despite their importance, the level of metadata reported in published studies varies substantially, and no community-agreed reporting standard currently exists for biomedical Raman spectroscopy. The development of a MIRRABI-style framework (Minimum Information Required for Reproducible Raman-Based Investigations) would provide a practical reference for authors, reviewers, and data repositories (Barton et al., 2022).

Wavenumber and intensity calibration represent additional reproducibility requirements that are inconsistently reported in biomedical Raman studies. These procedures ensure that Raman peak positions and relative intensities remain comparable across instruments and acquisition sessions (Ntziouni et al., 2022). While calibration is standard practice in metrological applications of Raman spectroscopy, it is less consistently documented in machine-learning studies. As a consequence, models trained on uncalibrated data may inadvertently learn instrument-specific artefacts rather than biologically meaningful spectral signatures, limiting their generalizability across laboratories and devices.

### Acquisition Protocols

Standardization of acquisition protocols concerns the procedures used to collect spectra from biological samples, including sample preparation, measurement geometry, and quality-control procedures. These choices can substantially influence both spectral quality and the biological information captured in the data. For example, tissue fixation methods are known to alter Raman signatures, while substrate selection may affect fluorescence background and the accessible spectral range (Wahl et al., 2022). Consequently, differences in acquisition protocols can introduce systematic variability that may be mistaken for biological signals if not carefully documented and controlled.

Measurement protocol decisions also influence the statistical properties of Raman datasets and the interpretation of machine-learning results. Choices such as spectral averaging, region-of-interest selection, exclusion of low-quality measurements, and quality-control procedures directly affect the data used for model development. However, these decisions are often insufficiently documented to allow independent replication. Although widely cited protocol papers provide guidance for biological Raman spectroscopy (Butler et al., 2016), they predate the current emphasis on machine-learning reproducibility and do not specifically address the requirements for constructing machine-learning-ready datasets. Developing community guidelines for Raman data collection, quality control, metadata recording, and dataset curation would therefore fill an important methodological gap.

Inter-laboratory and inter-instrument reproducibility remain major challenges for biomedical Raman spectroscopy. Differences in instrument configuration, calibration, and acquisition procedures can introduce systematic variability that limits the transferability of machine-learning models across sites. Various harmonization strategies, including calibration transfer and chemometric correction methods, have been proposed to improve comparability between instruments (Barton et al., 2022). Although these approaches have shown promise, their application to heterogeneous biomedical samples remains considerably more challenging than in industrial or pharmaceutical settings, highlighting the need for further standardization and validation efforts.

## Data Preprocessing

Preprocessing is one of the most influential stages of the biomedical Raman machine-learning pipeline because Raman spectra are affected not only by biological variation but also by numerous physical and instrumental factors, including fluorescence background, detector noise, cosmic ray artefacts, laser-power fluctuations, focus differences, substrate effects, and instrument-specific spectral distortions (Boateng, 2025). Effective preprocessing is therefore essential for separating biologically meaningful information from measurement-related variability. Common preprocessing workflows include quality control, calibration, baseline correction, denoising, normalization, and correction of scattering effects. Methods such as extended multiplicative signal correction (EMSC) are particularly valuable because they explicitly address unwanted physical variation in vibrational spectra (Afseth and Kohler, 2012). However, preprocessing should not be viewed as a purely technical step; it fundamentally shapes the spectral representation presented to downstream machine-learning models and can therefore influence both predictive performance and biological interpretation.

Preprocessing involves an important trade-off between noise reduction and preservation of biological signals. Aggressive smoothing, baseline correction, or dimensionality reduction may improve apparent classification performance while suppressing subtle biochemical differences associated with disease state, cellular phenotype, or tissue microenvironment. Conversely, insufficient preprocessing may allow models to learn instrument-, batch-, or sample-specific artefacts rather than biologically meaningful structure (Blake et al., 2022). Robust evaluation therefore requires sensitivity analyses across alternative preprocessing strategies, together with external validation and patient- or source-level data splitting. Particular care must be taken to avoid data leakage: preprocessing parameters should always be estimated exclusively from training data within each resampling fold, as performing normalization, feature selection, or dimensionality reduction before data splitting can substantially inflate performance estimates (Király and Tóth, 2025).

The importance of preprocessing is further amplified by its role as a major source of non-reproducibility in biomedical Raman machine-learning studies. As discussed earlier in this review, preprocessing choices, including baseline correction, denoising, normalization, spectral cropping, and dimensionality reduction, can substantially alter spectral representations and downstream model performance (Boateng, 2025). Consequently, two studies applying nominally similar preprocessing procedures may produce statistically incompatible results if they differ in parameter settings, processing order, or implementation details.

The problem is compounded by the fact that preprocessing procedures are often reported only at a high level. Algorithms for smoothing, baseline correction, normalization, and spectral selection are frequently described without the parameter settings required for independent reproduction. In addition, the rationale for specific preprocessing choices is rarely discussed, and differences between training, validation, and deployment pipelines are not always explicitly documented. As a result, studies that appear methodologically similar may in practice operate on substantially different spectral representations, making independent replication difficult even when the same raw data are available.

Reproducible preprocessing requires transparent reporting of the complete analytical pipeline, including processing order, parameter settings, software versions, quality-control procedures, and any spectral exclusions applied during analysis. When preprocessing involves learned components, such as PCA transformations or data-driven correction models, the corresponding parameters should be released alongside the code to ensure that new data can be processed consistently with the training data. Although this level of transparency is increasingly expected in computational biology and open-science initiatives, it has not yet become routine in the biomedical Raman literature (Blake et al., 2022).

Looking forward, community-maintained reference preprocessing pipelines for common biomedical Raman applications would provide both a practical resource for researchers and a benchmark against which new methods could be evaluated. Similar frameworks have played an important role in improving reproducibility in fields such as transcriptomics and single-cell genomics. Such pipelines should be openly available, versioned, and accompanied by reference datasets that allow independent verification of implementation and performance.  RamanSPy (Georgiev et al., 2024) represents an important step in this direction, but meaningful standardization will ultimately require coordinated efforts from researchers, instrument manufacturers, journals, funding bodies, and clinical stakeholders.

# Generalization and Validation

## Data Availability, Generalization, and Deployment

Despite rapid methodological advances, several practical barriers continue to limit the development and clinical translation of machine-learning models for biomedical Raman spectroscopy. These challenges include the scarcity of large, well-annotated datasets, variability across instruments and laboratories, limited availability of reproducible computational resources, and the difficulty of deploying models in real-world clinical environments. Addressing these issues is essential for developing Raman-based machine-learning systems that generalize beyond individual studies and operate reliably in routine biomedical practice.

A fundamental bottleneck in biomedical Raman machine learning is the scarcity of large, standardized, and publicly accessible datasets. Unlike computer vision and natural language processing, where millions of labelled examples are routinely available, most Raman studies rely on datasets containing only tens to hundreds of biological samples (Boateng, 2025). This reflects the practical realities of biomedical data collection: acquiring high-quality Raman spectra from well-characterized clinical material is time-consuming, expensive, and often subject to ethical and regulatory constraints (Qi et al., 2023). Consequently, complex machine-learning models, particularly deep-learning approaches, face a substantial risk of overfitting, while performance estimates obtained through internal validation may overstate true clinical utility (Beleites et al., 2013).

Compounding the scarcity problem is the pronounced heterogeneity across datasets collected in different laboratories. Raman spectra are highly sensitive to the conditions under which they are acquired: laser excitation wavelength, power, spot size, integration time, detector sensitivity, spectrometer resolution, objective magnification, and sample preparation protocol all introduce systematic differences specific to the instrument and laboratory (Boateng, 2025). As a result, a model trained on spectra acquired with one instrument setup frequently fails to generalize to another, even when the underlying biological classes are nominally identical. This instrument-to-instrument and laboratory-to-laboratory variability has been extensively noted as a barrier to cross-study comparison and evidence aggregation (Qi et al., 2023). While batch-correction and domain-adaptation strategies such as EMSC and transfer learning have been proposed to mitigate these effects (Afseth and Kohler, 2012), their effectiveness depends on the availability of paired or overlapping reference data, which is rarely provided in published studies. The absence of community-agreed acquisition standards and benchmark reference datasets further limits the extent to which reported results can be rigorously compared or reproduced (Blake et al., 2022).

## Label Granularity and the Challenge of Spectrum-Level Annotation

A more fundamental challenge concerns the mismatch between the granularity of biological labels and that of Raman measurements. In many studies, labels such as cancer subtype, drug-resistance status, or disease stage are assigned at the level of a cell line, tissue specimen, or patient, whereas spectra are acquired at the level of individual cells, pixels, or measurement locations. Consequently, all spectra originating from a given source often inherit the same label despite underlying biochemical heterogeneity. This is particularly problematic in tissue samples, where tumour, stromal, necrotic, and immune-cell populations may coexist within the same specimen (Blake et al., 2022).

This mismatch between label granularity and measurement granularity introduces a form of label noise that is difficult to quantify and correct. Models may partially learn sample-specific characteristics, such as growth conditions, passage number, inter-patient variation, or spatial sampling bias, rather than the biological phenomenon represented by the label (Luo et al., 2022). The problem becomes particularly severe when only a small number of biological sources are available, as models may effectively memorize source-specific spectral signatures instead of learning generalizable biological patterns. Under these conditions, cross-validation schemes that split individual spectra rather than biological sources can produce substantially inflated performance estimates. This form of data leakage is well recognized in chemometrics but remains common in the Raman machine-learning literature (Blake et al., 2022).

Addressing this challenge requires improvements in both data annotation and machine-learning methodology. On the data side, more detailed labels may be obtained through spatially resolved pathological annotation, single-cell phenotyping, or other approaches that better align biological labels with measurement locations. On the modelling side, weakly supervised and multiple-instance learning frameworks provide a natural way to represent situations in which labels are available only at the sample level rather than for individual spectra. Regardless of the learning strategy, strict patient- or source-level data splitting should be considered a minimum requirement for obtaining realistic performance estimates and avoiding inflation due to intra-source correlation (Beleites et al., 2013).

The challenges discussed throughout this section collectively define the pathway from promising machine-learning models to clinically deployable Raman-AI systems. Table 3 summarizes the principal barriers and representative mitigation strategies identified across the Raman machine-learning pipeline.

**Table 3. Translational barriers and recommended practices for clinically deployable machine learning in biomedical Raman spectroscopy.** Successful clinical deployment requires progress across the entire analytical pipeline, from data acquisition and preprocessing to model development, validation, explainability, reproducibility, and clinical integration. The table summarizes major barriers discussed throughout this review together with representative strategies proposed to improve robustness, generalizability, and clinical utility.

| **Domain** | **Major Challenge** | **Impact on Deployment** | **Representative Mitigation Strategies** |
|---|---|---|---|
| Data acquisition | Small, heterogeneous, and single-center cohorts | Limited generalizability and increased risk of overfitting | Multi-center studies, prospective recruitment, standardized cohort design |
| Instrumentation | Inter-instrument variability and acquisition-dependent spectral differences | Reduced transferability across laboratories and devices | Calibration transfer, harmonized acquisition protocols, instrument benchmarking |
| Preprocessing | Inconsistent preprocessing pipelines and reporting practices | Poor reproducibility and difficult comparison between studies | Standardized workflows, benchmark datasets, transparent reporting |
| Model development | Overfitting, data leakage, and inadequate evaluation protocols | Inflated performance estimates and poor real-world performance | Patient-level partitioning, external testing, rigorous benchmarking |
| Explainability | Limited biological interpretation and clinician trust | Reduced clinical acceptance and regulatory confidence | Spectral attribution methods, biomarker validation, integration with domain expertise |
| Reproducibility | Limited sharing of code, models, metadata, and workflows | Difficult replication and independent verification | Open-source software, FAIR principles, reproducible computational pipelines |
| Validation | Insufficient external and prospective validation | Uncertain clinical utility and deployment readiness | Multi-site validation, independent cohorts, prospective clinical studies |
| Clinical integration | Regulatory, workflow, and interoperability barriers | Delayed adoption in routine clinical practice | Real-time systems, interoperability standards, regulatory-grade validation |

## Clinical Translation and Real-time Deployment

While substantial progress has been achieved in machine-learning analysis of biomedical Raman spectra, routine clinical deployment will likely require coordinated advances in study design, standardization, validation, explainability, and data sharing. Table 4 summarizes the principal barriers identified throughout this review together with representative strategies that may facilitate robust and clinically deployable Raman-AI systems.

**Table 4**. Recommended practices for robust and clinically deployable Raman-AI systems.

| **Domain** | **Current Limitation** | **Recommended Practice** |
|---|---|---|
| Data acquisition | Small single-center cohorts | Multi-center collection and prospective studies |
| Instrumentation | Platform-specific biases | Calibration transfer and harmonized protocols |
| Preprocessing | Inconsistent pipelines | Transparent reporting and benchmark workflows |
| Model development | Overfitting and information leakage | Patient-level splits and external testing |
| Explainability | Limited biological interpretation | Spectral attribution and mechanistic validation |
| Reproducibility | Missing code and data | Open-source workflows and FAIR principles |
| Clinical deployment | Regulatory uncertainty | Prospective validation and clinical integration studies |

A challenge for clinical translation is the need to deliver reliable results within time frames compatible with real-world decision making. In diagnostic, surgical, and point-of-care settings, high predictive accuracy alone is insufficient if the underlying workflow requires extensive preprocessing, manual intervention, or computationally expensive inference. Real-time and near-real-time applications require the entire analytical pipeline (from spectral acquisition and quality control to model inference and reporting) to operate rapidly, robustly, and with minimal user intervention. Consequently, clinical utility depends not only on predictive performance but also on the ability to provide actionable information within operational time constraints (Boateng, 2025).

Achieving real-time performance requires careful consideration of computational efficiency throughout the analytical pipeline. Although deep-learning models often achieve strong predictive performance, their clinical deployment must be balanced against inference latency, hardware requirements, model complexity, and robustness to spectral variability. In some applications, simpler models may offer a more favorable trade-off between accuracy, interpretability, validation effort, and deployment cost. Future studies should therefore report not only predictive metrics such as accuracy or AUC, but also computational characteristics including preprocessing time, inference time, memory requirements, and scalability. Importantly, computational efficiency should not be pursued at the expense of reproducibility or robustness: models intended for deployment must still be validated across independent cohorts, instruments, and acquisition settings (Semmelrock et al., 2025).

A useful future direction is the development of open benchmark pipelines for real-time Raman classification. Such benchmarks should combine standardized datasets, reference preprocessing workflows, baseline models with different computational requirements, and evaluation metrics that capture both predictive performance and computational cost. For translational applications, reporting classification accuracy alone is insufficient; clinically useful models must also be reproducible, calibrated, robust to domain shift, and compatible with the time constraints of their intended application. Establishing such evaluation frameworks would help distinguish methods that perform well in retrospective studies from those that are genuinely ready for deployment in biomedical and clinical environments.

# Acknowledgments

This study was supported by the Ministry of Research, Innovation and Digitalization through the National Recovery and Resilience Plan (PNRR) of Romania, Pillar III, Component C9/Investment no. 8 (I8) - PNRR–III-C9–2023–I8, contract no 760096, ID proiect – CF 68/23.05.2023.

It was also supported through the Core Program within the National Research, Development and Innovation Plan 2022-2027, carried out with the support of MRID, project no.  2302101 (SIA-PRO), contract no 7N/2022.

It also received support from the "Large Language Models for the European Union (LLMs4EU)", project no. 101198470, call DIGITAL-2024-AI-B-06-LANGUAGE, funded by the European Union. Views and opinions expressed are however those of the author(s) only and do not necessarily reflect those of the European Union or the European Commission. Neither the European Union nor the granting authority can be held responsible for them.

# Author contributions

Conceptualization: B.O., A.P., I.P., M.P.

Literature review and manuscript preparation: B.O., A.M.S., N.S., L.M.S., A.S., J.S., M.N., A.-M.P., O.V., and E.M.

Methodological input: B.O., J.S., A.P., and I.P.

Supervision: B.O., I.P. and M.P.

Writing, review and editing: all authors.

All authors have read and approved the final manuscript.

# Competing Interests

The authors have no competing interests to declare.